\documentclass{article}

\usepackage[preprint]{neurips_2021}
\usepackage{makecell}
\usepackage[utf8]{inputenc} % allow utf-8 input
\usepackage[T1]{fontenc}    % use 8-bit T1 fonts
\usepackage{hyperref}       % hyperlinks
\usepackage{url}            % simple URL typesetting
\usepackage{booktabs}       % professional-quality tables
\usepackage{amsfonts}       % blackboard math symbols
\usepackage{nicefrac}       % compact symbols for 1/2, etc.
\usepackage{microtype}      % microtypography
\usepackage{xcolor}         % colors
\usepackage{amsfonts,amsmath} 
\usepackage{multirow}
\usepackage{mathtools}
\DeclareMathOperator*{\minimize}{\text{minimize}}
\DeclareMathOperator*{\st}{\text{subject to}}

\definecolor{fuming_color}{rgb}{0.39, 0.87, 0.09}

\title{Algorithm to Compilation Co-design: An Integrated View of Neural Network Sparsity}

\author{%
    Fu-Ming Guo \\
    Fidelity Investments \\
  \texttt{fuming.guo@fmr.com} \\
  
  \And
  
   \And
   Austin Huang \\
    Fidelity Investments \\
  \texttt{austinh@alum.mit.edu} \\
  
}

\begin{document}

\maketitle

\begin{abstract}

Reducing computation cost, inference latency, and memory footprint of neural networks are frequently cited as research motivations for pruning and sparsity. However, operationalizing those benefits and understanding the end-to-end effect of algorithm design and regularization on the runtime execution is not often examined in depth.

Here we apply structured and unstructured pruning to attention weights of transformer blocks of the BERT language model, while also expanding block sparse representation (BSR) operations in the TVM compiler. Integration of BSR operations enables the TVM runtime execution to leverage structured pattern sparsity induced by model regularization. 

This integrated view of pruning algorithms enables us to study relationships between modeling decisions and their direct impact on sparsity-enhanced execution. Our main findings are: 1) we validate that performance benefits of structured sparsity block regularization must be enabled by the BSR augmentations to TVM, with 4x speedup relative to vanilla PyTorch and 2.2x speedup relative to standard TVM compilation (without expanded BSR support).  2) for BERT attention weights, the end-to-end optimal block sparsity shape in this CPU inference context is not a square block (as in \cite{gray2017gpu}) but rather a linear 32x1 block 3) the relationship between performance and block size / shape is is suggestive of how model regularization parameters interact with task scheduler optimizations resulting in the observed end-to-end performance.

\end{abstract}

\section{Introduction}

Capabilities of neural networks have accelerated in the last decade and that progress has been accompanied by a productive tension between two competing goals. One goal is to expand the boundaries of functionality and performance, which has been accompanied by increasing scale in data and compute. A second goal is for new capabilities to have broad impact and operationalization. This goal tends towards the opposite direction - shrinking down compute and data required to achieve a capability.

For example, expansion of data and compute has led to recent NLP advances showing how large language models have unprecedented generalization capabilities \citep{brown2020language, raffel2019exploring}. These models should enable new realtime human-model interactions and entirely novel model development process where capabilities are instantiated at inference time, or can be rapidly adapted using lightweight methods such as prefix tuning \citep{li2021prefix}. However computational cost is an impediment to the impact and adoption of such models. How do we make these models accessible for both small and large scale research and deployment? Could such models be used in conjunction with privacy-preserving AI which requires model computation on edge devices? How can these language models be embedded at low cost into human-in-the-loop interactions requiring realtime latency? 

One proposed answer to these questions has been the literature around sparsification and pruning of neural networks. Since the 1980s, we have known that it is usually possible to prune most parameters from trained neural networks without affecting accuracy \citep{lecun1990optimal}. \cite{han2015deep} reduced number of parameters of AlexNet \citep{krizhevsky2012imagenet} by $9\times$ and VGG \citep{simonyan2014very} by $13\times$ using connection pruning. The lottery ticket hypothesis was proposed by \cite{frankle2018lottery}, which observes that a subnetwork of randomly-initialized network can replace the original network with the same performance. \cite{chen2020lottery, chen2021lottery} demonstrate the core LTH observations remain generally relevent in transformer models for both computer vision and natural language processing.  Although current-generation CPUs and GPUs do not immediately benefit from sparsity, there is an active research area dedicated to writing libraries to
accelerate sparse neural networks on these platforms \citep{elsen2020fast} and next generation hardware has native sparsity support (e.g., the NVIDIA A100, GraphCore IPU, and Cerebras Wafer-Scale Engine).

Although pruning is often motivated by performance, algorithms are often studied in isolation separate from their consequences with respect to compilation and execution. 
However, interactions between model regularization choices, model compilation, and inference execution can have subtle-yet-critical effects on performance.
 
In this research, we implement both unstructured and structured sparsification of the attention weights of BERT alongside BSR sparsity optimizations in the TVM compiler \citep{chen2018tvm}. We show how algorithms and compiler optimizations interact at different levels of the abstraction stack to determine end-to-end performance.

\section{Methods}

\subsection{Structured Sparsification}

Following the conventional pruning formulation, we consider the following optimization problem \citep{han2015deep}:
\begin{align}\label{eq: prune}
    \begin{array}{ll}
\displaystyle \minimize_{\mathbf w }        &  \displaystyle \sum f( \mathbf w) + \lambda \| \mathbf w \|_p,
    \end{array}
\end{align}
where$\|\mathbf{w}\|$ denotes the parameters of a neural network model, $\|\mathbf{w}\|_p$ denotes the $\ell_p$ norm of $\mathbf{w}$ for $p \in \{0, 1\}$. Note that equation \ref{eq: prune} can be interpreted as the Lagrangian form of the problem:
\begin{align}\label{eq: lagrangian}
    \begin{array}{ll}
    \displaystyle \minimize_{\mathbf{w} \in \mathbb R^d}    & \displaystyle f_0(\mathbf w) \\
    \st & \| \mathbf{w} \|_p  \leq \tau, 
    \end{array}
\end{align}
where $f_0$ is the pruning loss, $p\in\{0,1\}$, and $\tau$ is the tolerance of nonzero weights. To obtain models with structured sparsity, we calculate our norm $\| \mathbf{w}\|_p$ in a structured group manner 
\begin{align}\label{eq: group}
    \|\mathbf{w}\|_p = \sum_{n=1}^{N}\sum_{b=1}^{B}
\|\mathbf{w}_{b,n}\|_p
\end{align}
A weight matrix or convolution kernel can be divided into blocks with sparsity determined by the outcome of the model optimization. Here $B$ is the block size and $N$ is the number of blocks that comprise the weight matrix or convolution kernel. 

In contrast to the standard (unstructured) $\ell1$ / lasso procedure, group sparsity regularizes towards sparsity within each block, leading to a smaller set of more common used intra-block patterns, at least in the regime where $B$ is sufficiently small.

% To obtain variations of models with structured sparsity, we follow the generic optimization target propsed by [cite Wen 2016]:

% \[
% E(W) = E_D(W) + \lambda R(W) \cdot \sum_{l=1}^L{\sum_{g=1}^{G_l} ||w^{(g)}||_g}}
% \]

% Where $l$ iterates over the $L$ layers in the network, while $g$ iterates over the $G_l$ weights within layer. In contrast to the standard (unstructured) $L1$ / lasso procedure, group sparsity regularizes towards sparsity within each weight parameter, leading to a smaller set of more common used weight patterns, at least in the regime where $G_l$ is sufficiently small.

% $||w^(g)}||_g$ is defined as [cite Wen 2016]:

% \[
% ||w^(g)}||_g = \sqrt{\sum_{i=1}^{|w^{(g)}|}  (w_i^{(g)})^2
% \]

\subsection{TVM Compiler Integration}

\cite{gray2017gpu, gale2020sparse} has shown the advantage of block sparsity in executing Transformer \citep{vaswani2017attention} models on GPU. \cite{zhang2021full, gale2020sparse} demonstrates that the compiler scheduling introduces tremendous inference speed up on neural network. 
We augmented the TVM compiler with the following additions to achieve inference speed up on sparse neural network:

\begin{itemize}
    \item  We expand support for Block Sparse Row (BSR) for use with attention kernels and fully connected layers. BSR reduces the sparse neural network memory footprint and speeds up inference. Acceleration of sparse neural networks depends on eliminating operations (e.g., element-wise matrix multiplication) on zeroed weights (through pruning) and reusing the sparsity structure-based operations. 
    
    \item To eliminate the operation on zeroed-out weights, we implement the element-wise matrix multiplication for the BSR format. Specifically, we represent BSR matrices as data values, indices, and indptr (index pointer). Through indices and indptr, TVM picks only the non-zero weight in the sparse attention kernel and executes element-wise multiplication with input tensor. The BSR format and sparse multiplication operator implementation follow SciPy \citep{virtanen2020scipy}. 
    \item The TVM task scheduler is able to reuse structure-based sparsity. The aforementioned indices and indptr of BSR representation intrinsically reflect the characteristics of sparse matrices. The BSR representations are stored in a task buffer together with corresponding operators in TVM. TVM analyzes the similarity of tasks in the buffer and optimize the execution of the tasks through an auto-scheduler. The analysis proceeds in the task searching stage, attending to different hardware specifications (e.g., number of cores, cache size, instruction set architecture (ISA), max memory per block, and max thread per block). If two tasks in the task buffer are the same, TVM treats them as identical and reuse them. If two tasks are similar, TVM schedules them adjacent in the execution path.
\end{itemize}

\begin{figure}[ht]
    \centering
    \includegraphics[width=13.5cm]{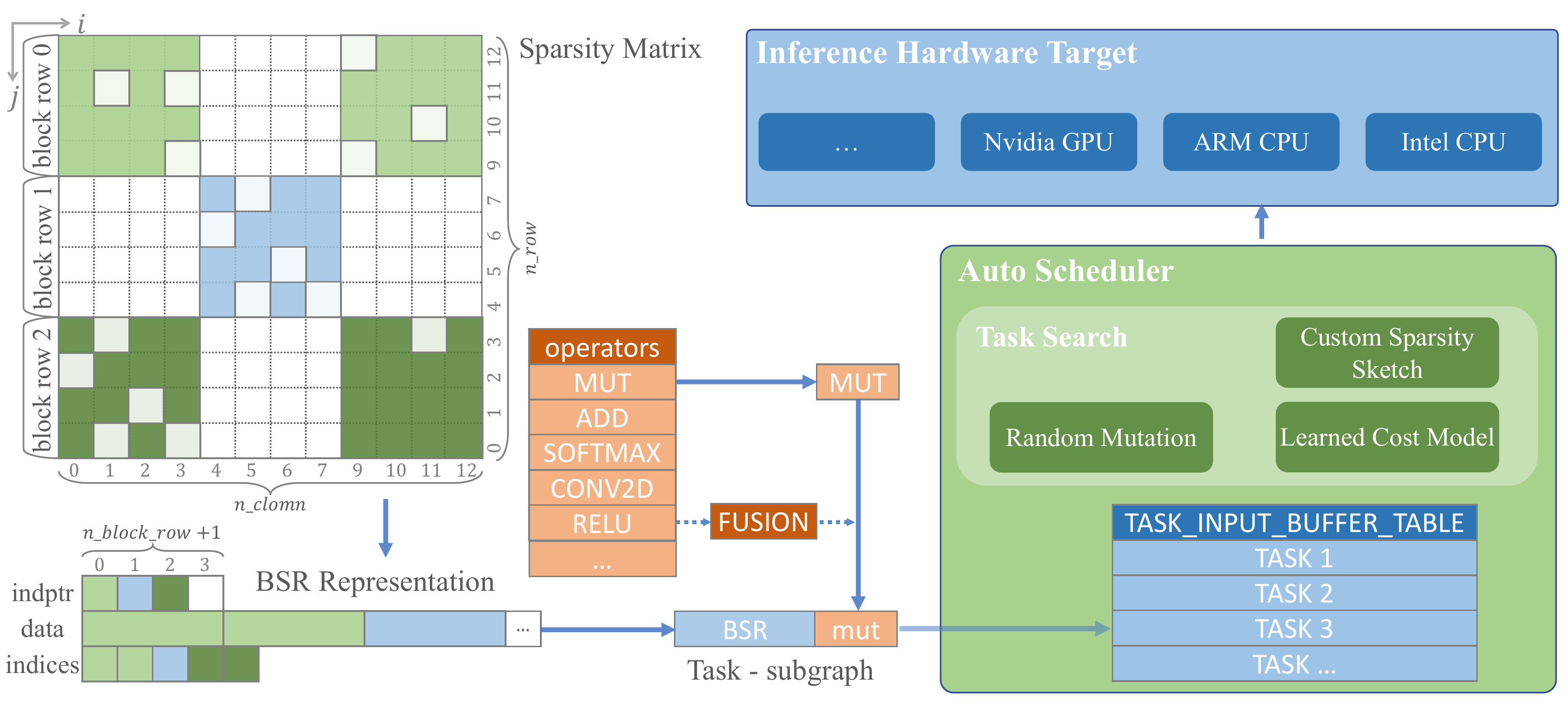}
    \caption{Overview of the augmented compiler: algorithm to compilation co-design}
    \label{fig:overview}
\end{figure}

\subsection{Experiments}

The goal of our experiments was to implement both unstructured and structured sparsification and assess the relative end-to-end impact on performance when the algorithm interacts with the compiler implementation, while assessing accuracy at different levels of sparsification.
Here we focus on a sparsity and compiler co-design approach to achieve inference speed up on CPU (Intel Core Processor Haswell). Haswell is not a high performance CPU, but rather a cost-effective contemporary standard and widely used in cloud computing environments. 

We use the official BERT model from Google as the starting point. Following the notation from \cite{devlin-etal-2019-bert}, we denote the number of layers (i.e., transformer blocks) as $L$, the hidden size as $H$, and the number of self-attention heads as $A$. We prune   the BERT model: $\mathrm { BERT } _ { \mathrm { BASE } }$ ($L=12, H=768, A=12, \text{total parameters}=110\mathrm{M}$). As the parameters of these transformer blocks take up more than $90\%$ weights of the entire BERT, the weights of these transformer blocks are our pruning target.

\textbf{Data:} In pre-training, we use the same pre-training corpora as \cite{devlin-etal-2019-bert}: BookCorpus ($800\mathrm{M}$ words) \citep{zhu2015aligning} and English Wikipedia ($2,500\mathrm{M}$ words). Based on the same corpora, we use the same preprocessing script\footnote{https://github.com/google-research/bert} to create the pre-training data. In fine-tuning, we report our results on the Stanford Question Answering Dataset (SQuAD) \citep{rajpurkar2016squad} and the General Language Understanding Evaluation (GLUE) benchmark \citep{wang2018glue}. The GLUE is a collection of datasets/tasks for evaluating natural language understanding systems\footnote{The datasets/tasks are: CoLA \citep{warstadt2018neural}, Stanford Sentiment Treebank (SST) \citep{socher2013recursive}, Microsoft Research Paragraph Corpus (MRPC) \citep{dolan2005automatically}, Semantic Texual Similarity Benchmark (STS) \citep{agirre2007semeval}, Quora Question Pairs (QQP), Multi-Genre NLI (MNLI) \citep{williams2017broad}, Question NLI (QNLI) \citep{rajpurkar2016squad}, Recognizing Textual Entailment (RTE) and Winograd NLI(WNLI) \citep{levesque2012winograd}.}.

\textbf{Input/Output representations:} We follow the input/output representation setting from \cite{devlin-etal-2019-bert} for both pre-training and fine-tuning. We use the WordPiece \cite{wu2016google} embeddings with a $30,000$ token vocabulary. The first token of every sentence is always a special classification token ([CLS]). The sentences are differentiated with a special token ([SEP]).

\textbf{Evaluation:} In pre-training, BERT considers two objectives: masked language modeling (MLM) and next sentence prediction (NSP). For MLM, a random sample of the tokens in the input sequence is selected and replaced with the special token $([\text{MASK}])$. The MLM objective is a cross-entropy loss on predicting the masked tokens. NSP is a binary classification loss for predicting whether two segments follow each other in the original text. In pre-training, we use MLM and NSP as training objectives to pre-train, retrain the BERT model, and as metrics to evaluate the BERT model . In fine-tuning, F1 scores are reported for SQuAD, QQP and MRPC. Matthew's Corr and Pearson-Spearman Corr are reported for CoLA and SST2 respectively. Accuracy scores are reported for the other tasks.

% \subsubsection{Inference speedup}

\section{Results}

We first run standard dense computations to set baselines for uncompiled PyTorch / Tensorflow inference (Table \ref{tab:inference}) without any model pruning or TVM compilation. We compare these baselines against sparse model variants using irregular sparse pruning and structured sparse pruning to BERT for a standard SQuAD QA task \citep{rajpurkar2016squad} and GLUE benchmark \citep{wang2018glue}. 

Next, with standard TVM compilation (i.e. prior to adding expanded BSR support) we observe an expected speedup, with inference time reduced to about 55\% of the vanilla PyTorch inference time (from 1389ms to 764ms). Note we are less interested in the absolute inference times which will be specific to a hardware configuration, and more interested in relative reduction observed in this context of commodity CPU hardware. Table \ref{tab:inference} shows inference times at an 80\% sparsity ratio for a range of block sparsity optimizations.

As a negative control, we apply irregular sparse pruning and structured sparse pruning of various dimensions and assess the inference time using the standard (unmodified) TVM compiler and runtime. We observe that inference performance remains approximately the same with most deviations being within $\sim$ 5\% of dense inference in spite of the 80\% sparsity ratio.

We then apply the same experiments for the augmented TVM compiler, expanding BSR support, labeled TVM$^+$ in Table \ref{tab:inference}. Here we observe notable performance improvements from the structured sparse pruning, with inferences times improving by as much as 55 \% (0.45 TVM$^+$/Dense) in the case of $1 \times 32$ block sparsity.

\begin{figure}[ht]
    \centering  
    \includegraphics[width=13.5cm]{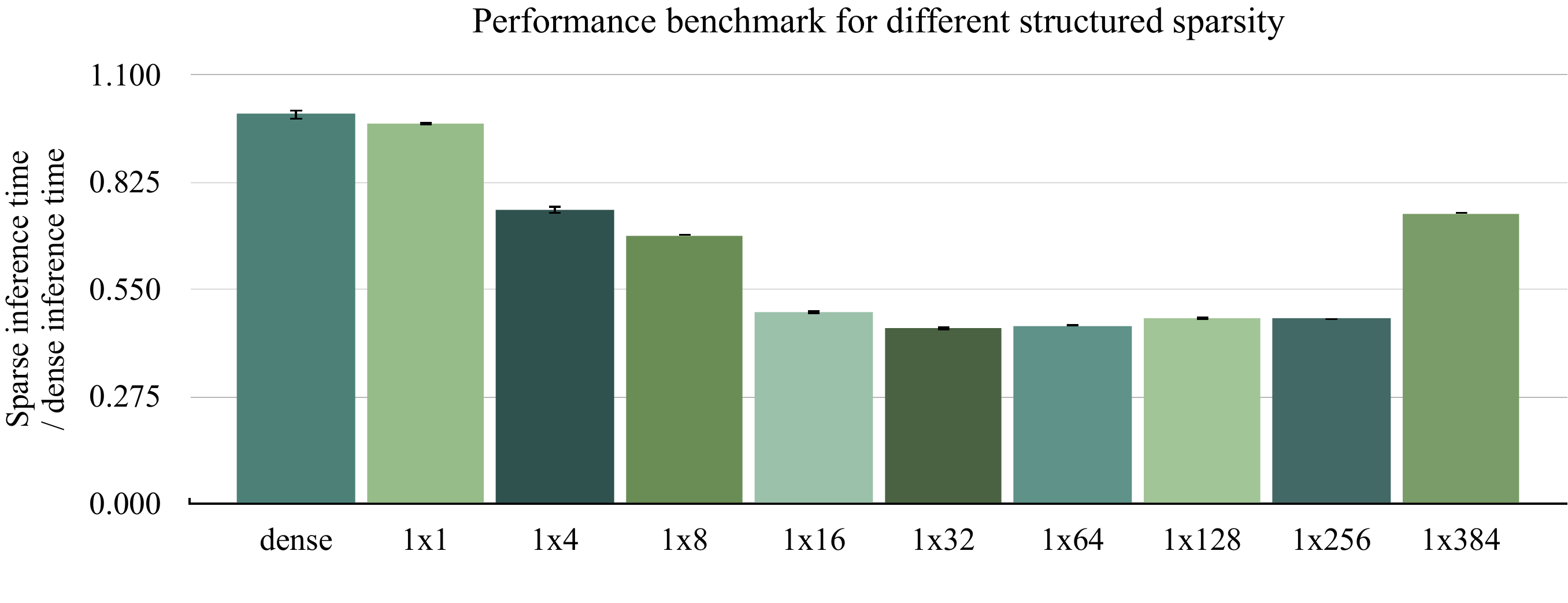}
    \caption{Performance benchmark for different structured sparsity}
    \label{fig:performance}
\end{figure}

There is a non-monotonic relationship between linear block size and computation time. Inference time improves for L1 block sparsity dimensions from $1 \times 1$ to $1 \times 32$, but becomes worse for larger sizes (Figure \ref{fig:performance}, Table \ref{tab:inference}).

When transitioning from single-row linear blocks for structured sparsity to square blocks of 4x4, 8x8, 16x16, and 64x64 we see a marked decrease in performance, although inference is still more performant than the dense computation.

\begin{center}
\begin{table}[ht]
\centering
\begin{tabular}{lllllll}
\toprule
\centering
                   & \makecell{$\ell_1$ \\ block size} & \makecell{PyTorch\\ms} & \makecell{Tensorflow\\ms} & \makecell{TVM ms\\mean std} & \makecell{TVM$^+$ ms\\mean std}& \makecell{TVM$^+$/Dense\\mean std}   \\
                   \hline
\centering
\makecell{Dense}              &       & 1389 & 1298    & 764 (19)   & 772 (19)      & 1.000 (0.025) \\
\hline
\centering
\makecell{Irregular\\Sparsity}   & 1 $\times$ 1   & 1375 & 1281    & 759 (14)    & 754 (6)       & 0.977 (0.008) \\
\hline
\makecell{Structured\\Sparsity} & 1 $\times$ 4   &         &            & 756 (28)     & 583 (17)      & 0.755 (0.022) \\
                   & 1 $\times$ 8   &         &            & 755 (11)     & 533 (2)       & 0.690 (0.003) \\
                   & 1 $\times$ 16  &         &            & 795 (13)    & 379 (8)      & 0.491 (0.010) \\
                   & 1 $\times$ 32  &         &            & 795 (9)     & 348 (5)       & 0.451 (0.006) \\
                   & 1 $\times$ 64  &         &            & 790 (10)    & 353 (5)       & 0.457 (0.006) \\
                   & 1 $\times$ 128 &         &            & 793 (12)    & 366 (8)     & 0.474 (0.010) \\
                   & 1 $\times$ 256 &         &            & 799(18)     & 366 (6)      & 0.474 (0.008) \\
                   & 1 $\times$ 384 &         &            & 779 (12)    & 576 (6)       & 0.746 (0.008) \\
                   & 4 $\times$ 4   &         &            & 751 (10)     & 556 (7)       & 0.720 (0.009) \\
                   & 8 $\times$ 8   &         &            & 776 (14)    & 529 (15)      & 0.685 (0.019)  \\
                   & 16 $\times$ 16 &         &            & 768 (6)     & 417 (6)       & 0.540 (0.008) \\
                   & 32 $\times$ 32 &         &            & 781 (9)     & 425 (4)       & 0.551 (0.005)  \\
                   & 64 $\times$ 64 &         &            & 760 (16)    & 427 (15)      & 0.553 (0.019) \\
                   \bottomrule \\
\end{tabular}
\caption{Inference times in milliseconds on a commodity Haswell Intel Core Processor and improvement relative to the dense baseline (TVM$^+$/Dense).}
\label{tab:inference}
\end{table}
\end{center}

The performance of the models (Table \ref{tab:accuracy}) is relatively consistent as the sparsity ratio varied from 50\% to 80\%. The largest drop was in SQuAD F1 scores while other tasks were within 1-3\% for 2x-5x compression rates. Fine tuning and hyperparameters were not aggressively optimized and these values can likely be improved with more computationally intensive fine tuning assessments.

\begin{table}[ht]
\tabcolsep=0.15cm
\centering
\begin{tabular}{rrlllllllll}
\toprule
\multicolumn{1}{l}{\makecell{Sparsity\\Ratio}} & {SQuAD 1.1} & {MNLI} & {MNLIM} & {MRPC} & {QNLI} & {QQP} & {RTE} & {SST-2} & {CoLA} \\
\hline
\makecell{Dense}                                                                         & 88.5                 & 84.1          & 84.1           & 84.6          & 91.4          & 90.4         & 69.7         & 93.2           & 81.5          \\
50\% Zeros                                                                            & 86.5                 & 83.6          & 82.5           & 87.0          & 90.9          & 89.7         & 68.6         & 92.1           & 81.9          \\
80\% Zeros                                                                            & 81.8                 & 81.3          & 81.1           & 86.8          & 89.5          & 89.0         & 64.6         & 91.5           & 80.4         \\
\bottomrule
\\
\end{tabular}
\caption{Task accuracy for dense, 50\% sparsified, and 80\% sparsified BERT variants.}
\label{tab:accuracy}
\end{table}

\section{Discussion}

We have implemented pruning algorithms alongside compiler support for sparse inference. In doing so, we are able to assess sparsity regularization from an integrated perspective. We show how the performance implications of modeling decisions are dependent on compiler support for sparsity, and illustrate the interaction between regularization and scheduling algorithms embedded in the compiler and runtime.

An interesting observation was the non-monotonic relationship between inference time and block size. This could reflect how modeling choices interact with the runtime scheduler. For small sparse blocks, block computation improves performance while the sparsity pattern is also likely to be replicated, leading to computation reuse by the TVM scheduler. As block sizes increase, in spite of larger-scale parallel computation, the cardinality of repeated sparsity patterns drops, which reduces the compute savings available to the scheduler. Thus larger linear patterns such as $1 \times 384$ as well as larger $N \times N$ square blocks are likely to have this issue. 

Some direct follow-ups to this work include 1) create instrumentation tools for introspection of task reuse by the scheduler to better quantify effects of regularization choices 2) examine whether the finding that $1 \times 32$ linear blocks are optimal relates to the sparsity patterns and structure of BERT's attention weight matrices 3) examine algorithm-to-compiler performance relationships in other architectures such as convolution and graph neural networks and 4) generalize principles for designing structured sparsification algorithms that are likely to result in end-to-end performance improvements, accounting for compilation and runtime execution.

\begin{ack}

Both authors are employeed by Fidelity Investments personal investing. They have no conflicts of interest to disclose.

\end{ack}

\bibliography{references}

\end{document}